\pdfoutput=1

\documentclass[11pt]{article}

\usepackage[]{EMNLP2023}

\usepackage{times}
\usepackage{latexsym}

\usepackage[T1]{fontenc}

\usepackage[utf8]{inputenc}

\usepackage{microtype}

\usepackage{inconsolata}
\usepackage{booktabs}
\usepackage{enumitem}
\usepackage{fdsymbol}
\usepackage{graphicx}
\usepackage{mathtools}
\usepackage{microtype}
\usepackage{multirow}
\usepackage{subfigure}
\usepackage{xspace}

\newcommand{\our}{\mbox{\textsc{SelfOOD}}\xspace}
\newcommand{\smallsection}[1]{\noindent\textbf{#1.}}

\newcommand\blfootnote[1]{%
  \begingroup
  \renewcommand\thefootnote{}\footnote{#1}%
  \addtocounter{footnote}{-1}%
  \endgroup
}

\title{\our: Self-Supervised Out-Of-Distribution Detection via Learning to Rank}

\author{
  Dheeraj Mekala$^{\spadesuit, \diamondsuit}$ $\quad$ Adithya Samavedhi$^{\spadesuit, \diamondsuit}$ $\quad$ Chengyu Dong$^{\diamondsuit}$ $\quad$  Jingbo Shang$^{\diamondsuit, \heartsuit, *}$ \\
  $^\diamondsuit$University of California San Diego\\
  $^\heartsuit$ Hal\i c\i o\u glu Data Science Institute, University of California San Diego\\
  \small \texttt{\{dmekala, asamavedhi, cdong, jshang\}@ucsd.edu}
}

\begin{document}
\maketitle

\begin{abstract}
    \blfootnote{$\spadesuit$ Equal Contribution}
    \blfootnote{$*$ Jingbo Shang is the corresponding author.}
    Deep neural classifiers trained with cross-entropy loss (CE loss) often suffer from poor calibration, necessitating the task of out-of-distribution (OOD) detection.
    Traditional supervised OOD detection methods require expensive manual annotation of in-distribution and OOD samples. 
    To address the annotation bottleneck, we introduce \textbf{\our}, a \textbf{self}-supervised \textbf{OOD} detection method that requires only in-distribution samples as supervision.
    We cast OOD detection as an inter-document intra-label (IDIL) ranking problem and train the classifier with our pairwise ranking loss, referred to as IDIL loss.
    Specifically, given a set of in-distribution documents and their labels, for each label, we train the classifier to rank the softmax scores of documents belonging to that label to be higher than the scores of documents that belong to other labels.
    Unlike CE loss, our IDIL loss function reaches zero when the desired confidence ranking is achieved and gradients are backpropagated to decrease probabilities associated with incorrect labels rather than continuously increasing the probability of the correct label.
    Extensive experiments with several classifiers on multiple classification datasets demonstrate the effectiveness of our method in both coarse- and fine-grained settings.

\end{abstract}
\section{Introduction}
    Deep neural networks (DNN) are ubiquitously used for Text classification~\cite{liu2019roberta, Devlin2019BERTPO, Yang2019XLNetGA, NEURIPS2020_1457c0d6}.
    However, they are generally poorly calibrated, resulting in erroneously high-confidence scores for both in-distribution and out-of-distribution (OOD) samples~\cite{szegedy2013intriguing, nguyen2015deep, guo2017calibration, mekala-etal-2022-lops}.
    Such poor calibration makes DNNs unreliable and OOD detection task vital for the safe deployment of deep learning models in safety-critical applications~\cite{moon2020confidence}.


    \begin{figure}[t]
        \subfigure[CE Loss]{
            \includegraphics[width=\linewidth]{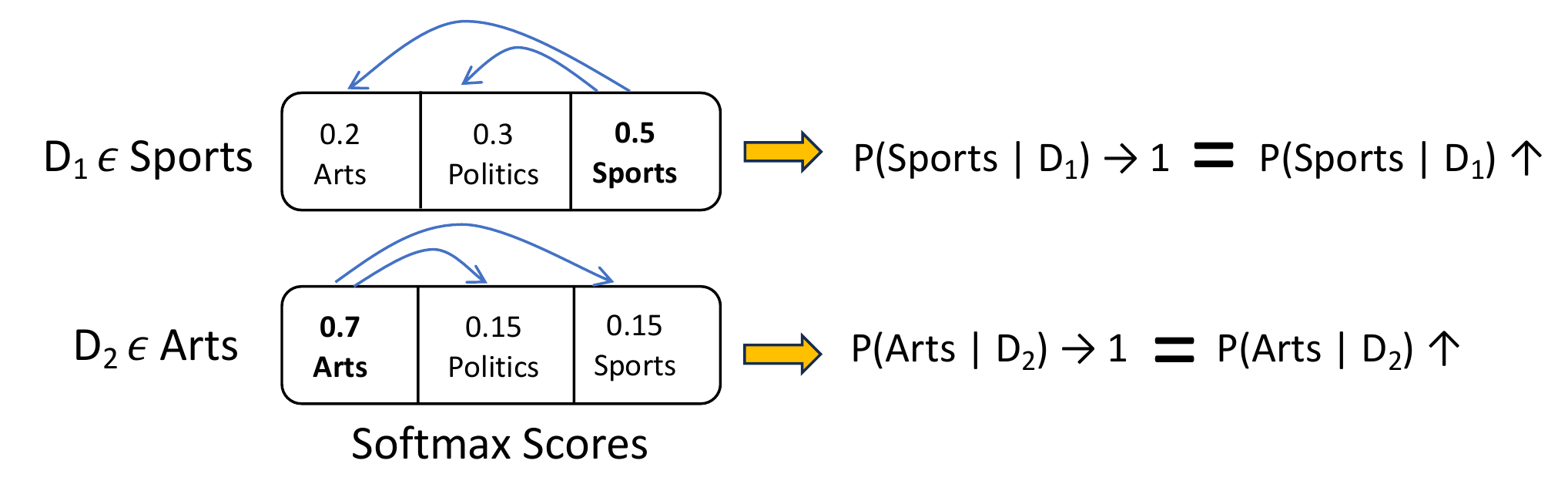}
            \label{fig:example_ce}
        }
        \subfigure[\our]{
            \includegraphics[width=\linewidth]{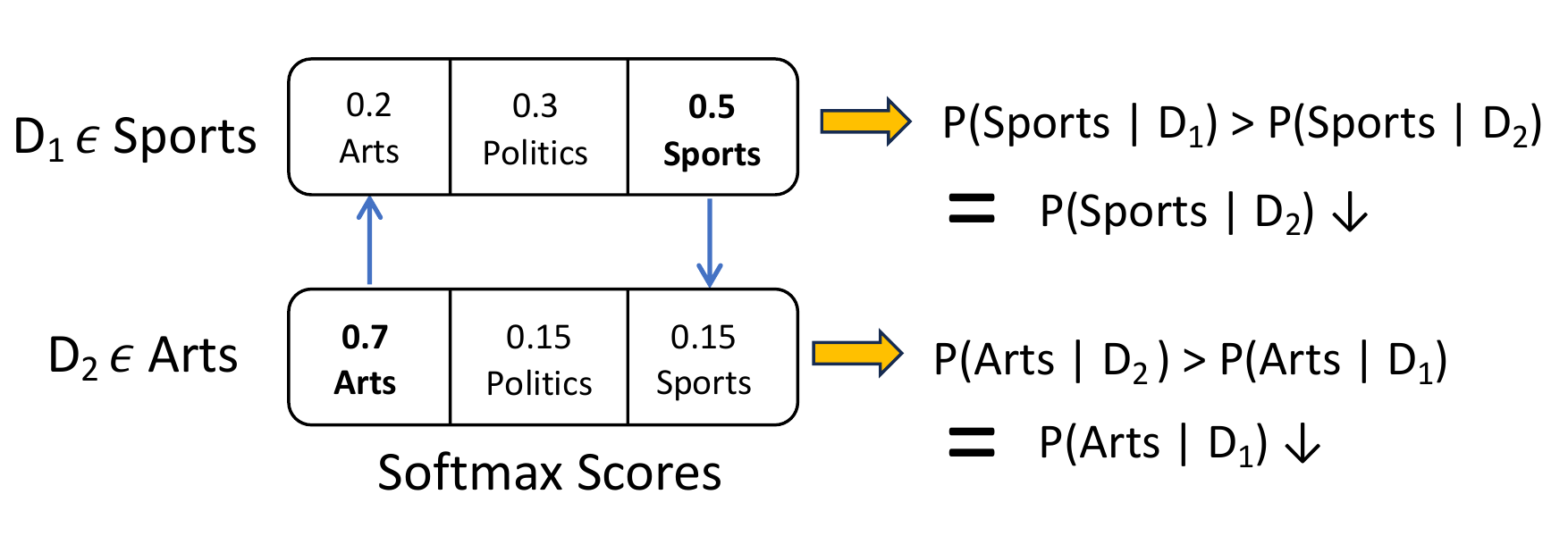}
            \label{fig:example_selfood}
        }
        \vspace{-3mm}
        \caption{CE Loss and \our optimization for two documents $D_1, D_2$ belonging to Sports and Arts. CE loss increases the scores corresponding to the Sports class for $D_1$ and Arts class for $D_2$, implying an intra-document comparison. Instead, \our compares the softmax scores in an inter-document intra-label fashion where it reduces the scores corresponding to Sports class for $D_2$ to be less than that of $D_1$.}
        \vspace{-5mm}
    \end{figure}

    \begin{figure*}[t]
        \center
        \includegraphics[width=\linewidth]{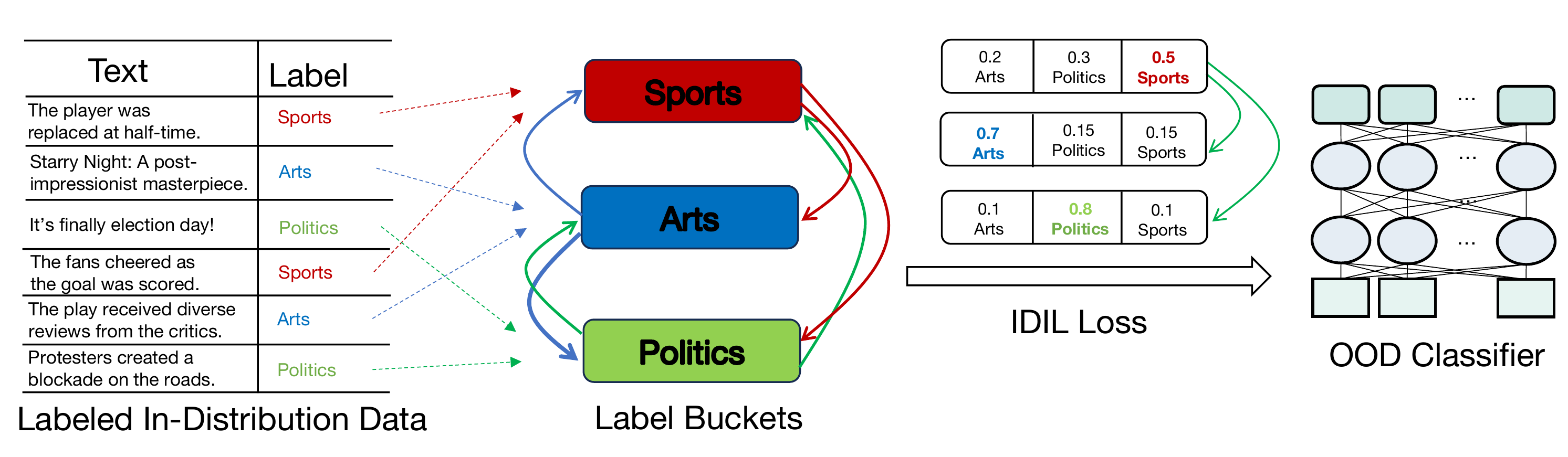}
        \caption{\our is a self-supervised framework that requires only annotated in-distribution data to train the OOD classifier. Firstly, we bucket documents based on their annotated label (in dotted lines). Then, we compare each document in a bucket with all documents in other buckets to compute IDIL loss (in solid lines). Finally, we backpropagate gradients to decrease scores associated with incorrect labels during the training of the OOD classifier.}
        \label{fig:overview}
    \end{figure*}
        
    Traditional supervised OOD detection methods \cite{hendrycks2018deep, larson2019evaluation, kamath2020selective, zeng2021adversarial} assume access to high-quality manually annotated in-distribution and OOD samples. However, this requires extensive annotation of OOD samples belonging to diverse distributions, which is expensive to obtain.
    Moreover, text classifiers are ideally desired to be more confident on in-distribution samples than OOD samples. 
    However, the poor calibration of DNN precludes this phenomenon.
    
    To address these problems, we propose \our, a self-supervised OOD detection framework that requires only in-distribution samples as supervision.
    Deep learning models are desired to be more confident in in-distribution samples than OOD samples.
    To adhere to this constraint, we formulate OOD detection as an inter-document intra-label (IDIL) ranking problem and train the classifier using our pairwise ranking loss, referred to as IDIL loss.
    As shown in Figure-\ref{fig:example_ce}, text classifiers are generally trained using cross-entropy loss (CE loss)~\cite{good1952rational} in an intra-document fashion where for each document, the classifier is trained to distinguish between different labels by maximizing the score corresponding to the correct label. 
    Instead, in our method, as shown in Figure~\ref{fig:example_selfood}, we propose to train in an inter-document, intra-label fashion where for each label, we train the model to rank the considered label probability score in documents belonging to the label to be higher compared to those not belonging to it.
    As OOD documents generally don't belong to any label, we hypothesize such explicit training to rank, translates to accurately distinguishing OOD from in-distribution documents.

    Moreover, minimizing CE loss involves continuous optimization to increase the probability of the correct label over the other labels, making the classifier overconfident~\cite{wei2022mitigating}.
    Instead, in our method, our IDIL loss function becomes zero once the desired ranking is achieved, and during training, we backpropagate gradients to decrease probabilities associated with incorrect labels rather than increasing the probability of the correct label.
    Theoretically, the perfect model trained using the CE loss is a solution to our ranking problem, however, empirically, we observe that our ranking objective leads to a different solution, demonstrating the importance of the optimization procedure.
    Finally, it is important to note that our ranking formulation, loss function, and the self-supervised training strategy have been specifically designed to improve the performance of OOD detection, rather than classification accuracy.

    We present our framework in Figure~\ref{fig:overview}.
    Given a set of in-distribution documents and corresponding labels as input, we bucket documents belonging to each label and train the classifier to rank the probabilities using our IDIL loss function. 
    Specifically, for each document in a label bucket, we pair up with all documents in every other bucket and compute IDIL loss.


    Our contributions are summarized as follows:
    \begin{itemize}[leftmargin=*,nosep]
        \item We propose \our, a novel self-supervised method to train an OOD detection model without any OOD samples as supervision.
        \item We formulate OOD detection as an inter-document intra-label ranking problem and optimize it using our IDIL ranking loss.
        \item We perform extensive experiments on multiple text classification datasets to demonstrate the effectiveness of our method in OOD detection.
        \item We release the code on Github\footnote{\url{https://github.com/dheeraj7596/SELFOOD}}.
    \end{itemize}
\section{Related Work}
Traditional supervised methods cast the OOD detection as classification with binary labels~\cite{kamath2020selective}, one additional label for unseen classes~\cite{fei2016breaking}.
Recent works have deviated from requiring extensive labeled OOD data to leverage various distance metrics to detect OOD samples.
\cite{leys2018detecting, xu2020deep} use Mahalanobis distance as a post-processing technique to identify OOD samples. 
These methods use distance-based scoring functions along with the intermediate model layer features to determine an OOD score. 
\cite{lee2018simple, hsu2020generalized} use a similar distance metric called ODIN to detect OOD images.
An alternate approach to compensate for the lack of OOD training data involves generating pseudo-OOD data for training. 
\cite{ouyang2021energy} propose a framework to generate high-quality OOD utterances and importance weights by selectively replacing phrases in in-domain samples. 
\cite{zhan2021out} generate pseudo-OOD samples for the task of intent detection using self-supervision.
\cite{zhou2021contrastive, zeng2021modeling} introduce self-supervised approaches to ood detection using a contrastive learning framework. 
They suggest fine-tuning transformers using a margin-based contrastive loss to learn text representations for OOD classification. 
\cite{vyas2018out, li2021k} treat a part of in-domain data as OOD samples as an alternate self-supervised approach. 
Further, \cite{wu2022revisit} use a Reassigned Contrastive Loss (RCL) along with an adaptive class-dependent threshold mechanism to separate in-domain and OOD intents. \cite{ren2019likelihood, gangal2020likelihood} leverage likelihood ratios crafted by generative models to classify OOD samples. 
\cite{wei2022mitigating} observe that the norm of the logits keeps increasing during training, leading to overconfident outputs, and propose LogitNorm as a solution to decoupling the output norm during training optimization. 
\cite{moon2020confidence} introduce a novel Correctness Ranking Loss function in order to regularize output probabilities to produce well-ranked confidence estimates.  
Other calibration techniques include "Top-label" calibration which is used to regularize the reported probability for the predicted class~\cite{gupta2021top}. 
\section{\our: Self-Supervised OOD Detection}
In this section, we first present the problem statement, then briefly discuss the motivation of our ranking formulation for OOD detection, and finally describe our method including the loss function and its optimization strategy.

\subsection{Problem Statement}
In this paper, we work on the Out-of-distribution detection task with only in-distribution samples as supervision. 
Specifically, given a labeled dataset $\mathcal{D}^{InD}=\{(x_1, y_1), (x_2, y_2), \ldots (x_n, y_n)\}$ sampled from a distribution space $(\mathcal{X}, \mathcal{C})$ where documents $x_i \in \mathcal{X}$ and labels $y_i \in \mathcal{C}$ as input,
our goal is to train an OOD detector $\mathcal{M}$ that accurately distinguishes in-distribution documents $\mathcal{D}^{InD}$ and OOD documents $\mathcal{D}^{OOD}\notin(\mathcal{X}, \mathcal{C})$ without any OOD documents required for training.

\subsection{Motivation}
Numerous neural text classifiers have been proposed, incorporating multiple hidden layers~\cite{rosenblatt1957perceptron}, convolutional layers~\cite{kim-2014-convolutional}, and various types of attention mechanisms~\cite{Devlin2019BERTPO, liu2019roberta, Radford2019LanguageMA}.
All these models culminate in a softmax head, which produces probabilities corresponding to each class.
These classifiers are generally trained with CE loss in an intra-document fashion i.e. each document is considered independently and the softmax score of the true label is maximized.
Such training of neural text classifiers is known to increase the magnitude of logit vectors even when most training examples are correctly classified~\cite{wei2022mitigating}, making them poorly calibrated that produce unreasonably high probabilities even for incorrect predictions~\cite{szegedy2013intriguing, nguyen2015deep, guo2017calibration}.
This diminishes their ability to maintain the desired attribute of ordinal ranking for predictions based on confidence levels, wherein a prediction exhibiting a higher confidence value should be considered more likely to be accurate than one with a lower confidence value~\cite{moon2020confidence}.
Intuitively, a text classifier possessing such quality would be a perfect OOD detector. 

\subsection{OOD Detection as Inter-Document Intra-Label Ranking}
In order to align with the aforementioned characteristic, we propose formulating the OOD detection as an inter-document intra-label ranking problem.
Specifically, given a set of in-distribution documents, we compare across documents but within the same label and train our model to generate higher probability score for documents belonging to the label than for documents not belonging to the label.
We consider the same model architecture as any text classifier with a softmax head that generates scores corresponding to each label, however, we train it using our IDIL loss instead of CE loss.
Our assumption is that an OOD document does not fall under any specific label in the in-distribution space. 
Hence, we anticipate that the trained model would produce lower scores for OOD documents compared to in-distribution documents. 
This distinction in scores is expected to facilitate easy separation between OOD and in-distribution documents.

\noindent\textbf{IDIL Loss} is a pairwise-ranking loss that enforces desired ordinal ranking of confidence estimates.
This loss function reaches its minimum value for a particular label when the probability of that label being the annotated label is greater than its probability when it is not the annotated label.
Specifically, for documents $x_1, x_2 \in \mathcal{D}^{InD}$ and their corresponding annotated labels $y_1, y_2$ where $y_1 \neq y_2$, IDIL loss corresponding to label $y_1$ is mathematically computed as follows:
\begin{equation}\label{eqn:idil}
    \resizebox{\hsize}{!}{$\mathcal{L}_{IDIL}(y_1 | x_1, x_2) = SiLU(p(y_1 | x_2) - p(y_1 | x_1))$}
\end{equation}
where $SiLU(x) = x \sigma(x)$ is the Sigmoid Linear Unit (SiLU) function~\cite{elfwing2018sigmoid}.
To ensure stable training and enhance performance, we incorporate the SiLU function, a continuous variant of the Rectified Linear Unit (ReLU)~\cite{hahnloser2000digital}, in conjunction with the ranking loss. 
The SiLU function introduces smooth gradients around zero, effectively mitigating potential instability issues during training. 
We observe that this inclusion contributes to the overall stability in training and improved performance of the model as shown in Section~\ref{sec:abl}.
Note that, in contrast to CE loss, IDIL loss becomes zero once the desired ranking is achieved, addressing the overconfidence issue.

\subsection{Implementation}
Ideally, the loss has to be computed over all possible pairs of documents for each model update.
However, it is computationally expensive.
Therefore, following ~\cite{toneva2018an, moon2020confidence}, we approximate the computation by considering only documents in each mini-batch.
Specifically, we bucket the documents in the mini-batch based on their annotated label and pair each document in a bucket with all documents in other buckets and compute the loss. 
Mathematically, the loss for a mini-batch $b$ is computed as follows:
\[
\mathcal{L} = \sum_{l \in \mathcal{C}} \sum_{x_1 \in b_l} \sum_{x_2 \in b_{\neg l}} \mathcal{L}_{IDIL}(l | x_1, x_2)
\]
where $b_l$ denotes the set of training data points $x$ in this batch $b$ whose label are $l$, and $b_{\neg l}$ denotes the set of training data points $x$ in this batch $b$ whose label is \emph{not} $l$.


In contrast to CE loss, where the optimization involves increasing the score corresponding to the correct label, we backpropagate gradients to decrease scores associated with incorrect labels.
Specifically, during the backpropagation of gradients, we detach the gradients for the subtrahend of the difference and exclusively propagate the gradients through the minuend. 
In Equation~\ref{eqn:idil}, for instance, we detach the gradients for $p(y_1 | x_1)$ and solely backpropagate the gradients through $p(y_1 | x_2)$. 
This detachment allows for a more controlled and selective gradient flow, aiding in the optimization process, and improvement in performance as shown in Section~\ref{sec:abl}.

It is important to note that our optimization focuses solely on the inter-document ranking loss. Consequently, while the trained model would serve as a reliable OOD detector, it may not perform as effectively as a classifier.
\section{Experiments}
\begin{table}[t]
    \center
    \vspace{-3mm}
    \scalebox{0.82}{
    \begin{tabular}{c c c c c}
        \toprule
            {\textbf{Dataset}} & {\textbf{Domain}} & {\textbf{Criteria}} &  {\textbf{\# Docs}} & {\textbf{\# labels}} \\
        \midrule
        \textbf{NYT} & News & Topic & 13081 & 26 \\
        \textbf{Yelp} & Reviews & Sentiment & 70000 & 5 \\
        \textbf{Clima} & Climate & Question Type & 17175 & 8 \\
        \textbf{TREC} & General & Question Type & 5952 & 6\\
        \bottomrule
    \end{tabular}
    }
    \caption{Dataset statistics.
    }
    \label{tbl:datastats}
\end{table}

\begin{table*}[t]
\centering
\small
\resizebox{\linewidth}{!}{
\begin{tabular}{c c c cccc cccc}
\toprule
& & & \multicolumn{4}{c}{BERT} & \multicolumn{4}{c}{RoBERTa} \\
\cmidrule(lr){4-7} \cmidrule(lr){8-11}
In-dist & OOD & Method &  FPR95($\downarrow$) & ERR($\downarrow$) & AUROC($\uparrow$) & AUPR($\uparrow$) &  FPR95($\downarrow$) & ERR($\downarrow$) & AUROC($\uparrow$) & AUPR($\uparrow$) \\
\midrule
\multirow{9}{*}{Yelp} & \multirow{2}{*}{NYT} & CE Loss & 82.1 & 34.9 & 63.4 & 50.7 & 79.8 & 32.9 & 66.8 & 58.0 \\
                      & & \our & \textbf{63.2} & \textbf{19.6} & \textbf{79.4} & \textbf{82.7} & \textbf{69.5} & \textbf{21.6} & \textbf{72.6} & \textbf{78.8} \\
                      & & CRL & 99.9 & 36.9 & 43.9 & 51.1 & 99.8 & 28.1 & 64.8 & 60.5\\
                      \cmidrule{2-11}
                      & \multirow{2}{*}{Clima} & CE Loss & 83.2 & 27.8 & 48.6 & 37.0 & 76.4 & 22.0 & \textbf{78.2} & 68.4 \\
                      & & \our & \textbf{17.6} & \textbf{4.8} & \textbf{97.9} & \textbf{96.4} & \textbf{65.6} & \textbf{16.6} & 73.2 & \textbf{77.2} \\
                      & & CRL & 99.6 & 29.0 & 49.1 & 38.8 & 99.9 & 20.6 & 63.0 & 56.9 \\
                      \cmidrule{2-11}
                      & \multirow{2}{*}{TREC} & CE Loss & 74.2 & 33.7 & 52.8 & 62.7 & 58.5 & 27.5 & 75.3 & 78.3 \\
                      & & \our & \textbf{0.0} & \textbf{0.0} & \textbf{100.0} & \textbf{100.0} & \textbf{55.1} & \textbf{20.6} & \textbf{80.2} & \textbf{90.3}\\
                      & & CRL & 33.3 & 15.5 & 66.7 & 77.2 & 100.0 & 37.3 & 56.7 & 73.9\\
                      \midrule
                      \multirow{9}{*}{NYT} & \multirow{2}{*}{Clima} & CE Loss & 22.1 & \textbf{1.0} & 96.9 & \textbf{92.5} & 15.7 & \textbf{1.0} & 97.3 & 92.8\\
                      & & \our & \textbf{10.5} & 1.6 & \textbf{98.4} & 90.5 & \textbf{4.6} & \textbf{1.0} & \textbf{99.3} & \textbf{95.4}\\
                      & & CRL & 32.3 & 1.7 & 95.4 & 86.5 & 79.3 & 3.2 & 79.8 & 67.0\\
                      \cmidrule{2-11}
                      & \multirow{2}{*}{TREC} & CE Loss & 14.5 & 5.8 & 97.0 & 87.1 & 21.9 & 6.7 & 95.5 & 83.5\\
                      & & \our & \textbf{0.0} & \textbf{0.0} & \textbf{100.0} & \textbf{100.0} & \textbf{0.0} & \textbf{0.0} & \textbf{100.0} & \textbf{100.0} \\
                      & & CRL & 67.9 & 6.5 & 89.9 & 79.1 & 85.6 & 9.6 & 71.4 & 61.4 \\
                      \cmidrule{2-11}
                      & \multirow{2}{*}{Yelp} & CE Loss & \textbf{3.4} & \textbf{0.7} & \textbf{99.2} & \textbf{82.3} & 1.8 & 0.6 & 99.3 & 85.4 \\
                      & & \our & 22.5 & 1.1 & 95.7 & 57.8 & \textbf{0.2} & \textbf{0.2} & \textbf{99.9} & \textbf{97.5} \\
                      & & CRL & 69.5 & 0.5 & 89.3 & 77.8 & 73.3 & 1.0 & 83.8 & 58.3 \\
                      \midrule
                      \multirow{9}{*}{Clima}  & \multirow{2}{*}{TREC} & CE Loss & 34.9 & 5.6 & 94.9  & 92.2 & 60.2 & 9.1 &  \textbf{90.4} & 83.8 \\
                      & & \our & \textbf{0.3} & \textbf{1.1} & \textbf{99.4} & \textbf{99.4} & \textbf{38.9} & \textbf{8.8} & 90.1 & \textbf{84.7} \\
                      & & CRL & 83.0 & 12.7 & 82.0 & 72.2 & 60.9 & 12.6 & 79.2 & 63.2 \\
                      \cmidrule{2-11}
                      & \multirow{2}{*}{Yelp} & CE Loss & 46.3 & 0.9 & 93.8 & 80.1 & 79.5 & 1.9 & \textbf{86.5} & 47.2 \\
                       & & \our & \textbf{14.7} & \textbf{0.6} & \textbf{98.0} & \textbf{90.8} & \textbf{37.6} & \textbf{1.5} & 72.3 & \textbf{54.9} \\
                      & & CRL & 98.7 & 1.4 & 67.3 & 53.3 & 72.9 & 2.1 & 50.9 & 19.0 \\
                      \cmidrule{2-11}
                       & \multirow{2}{*}{NYT} & CE Loss & 44.9 & 4.8 & 93.2 & 84.4 & 62.5 & \textbf{6.7} & \textbf{89.6} & \textbf{75.2} \\
                       & & \our & \textbf{17.2} & \textbf{2.6} & \textbf{97.8} & \textbf{94.5} & \textbf{55.3} & \textbf{6.7} & 68.6 & 61.5 \\
                      & & CRL & 85.6 & 7.6 & 78.0 & 62.5 & 61.3 & 9.4 & 61.8 & 39.9 \\
                      \midrule
                      \multirow{9}{*}{TREC}  & \multirow{2}{*}{Yelp} & CE Loss & 27.2 & 0.3 & 94.9 & 71.2 & 11.3 & 0.4 & 96.9 & 62.3 \\
                       & & \our & \textbf{0.0} & \textbf{0.0} & \textbf{100.0} & \textbf{100.0} & \textbf{0.0} & \textbf{0.0} & \textbf{100.0} & \textbf{100.0} \\
                       & & CRL & 96.6 & 0.7 & 70.9 & 16.5 & 99.8 & 0.7 & 24.3 & 9.7 \\
                      \cmidrule{2-11}
                      & \multirow{2}{*}{NYT} & CE Loss & 8.4 & 1.3 & 97.6 & 87.5 & 6.8 & 1.4 & 98.1 & 87.1 \\
                      & & \our & \textbf{0.0} & \textbf{0.0} & \textbf{100.0} & \textbf{100.0} & \textbf{0.0} & \textbf{0.0} & \textbf{100.0} & \textbf{100.0} \\
                     & & CRL & 97.2 & 3.3 & 73.0 & 35.2 & 99.9 & 3.8 & 22.9 & 12.2 \\
                      \cmidrule{2-11}
                      & \multirow{2}{*}{Clima} & CE Loss & 15.1 & 1.7 & 96.2 & 69.5 & 12.2 & 1.8 & 96.4 & 67.8 \\
                       & & \our & \textbf{0.0} & \textbf{0.0} & \textbf{100.0} & \textbf{100.0} & \textbf{0.0} & \textbf{0.0} & \textbf{100.0} & \textbf{100.0}\\
                      & & CRL & 86.3 & 2.2 & 80.7 & 39.5 & 98.9 & 2.7 & 30.1 & 13.0 \\

\bottomrule
\end{tabular}
}
\caption{OOD detection results with BERT \& RoBERTa classifiers. Each experiment is repeated with three random seeds and the mean scores are reported. The false-positive-rate at 95\% true-positive-rate (FPR95), minimum detection error over all thresholds (ERR), the area under the risk-coverage curve (AURC), and the area under the precision-recall curve (AUPR) using in-distribution samples as the positives are used as evaluation metrics.
}
\label{results-cls-all}
\end{table*}

We evaluate our OOD detection method against state-of-the-art baselines with two classifiers on multiple datasets belonging to different domains. 
In this section, we present our experimental settings, compared methods, and performance.

\subsection{Datasets}
We evaluate our method and baselines on four publicly available English text classification datasets belonging to different domains.
In particular, we consider the news topic classification dataset New York Times (NYT)\footnote{\url{http://developer.nytimes.com/}}, restaurant review sentiment classification dataset Yelp\footnote{\url{https://www.yelp.com/dataset/}}, and question-type classification datasets related to climate: Clima-Insurance+ (Clima)~\cite{laud2023climabench}, and a general domain: TREC~\cite{li-roth-2002-learning, hovy-etal-2001-toward}.
The documents within the New York Times dataset are labeled with both coarse and fine-grained labels. 
For our training and testing process, we utilize fine-grained labels.
The dataset statistics are provided in Table~\ref{tbl:datastats}.

\subsection{Compared Methods}
We compare with several OOD detection methods mentioned below:
\begin{itemize}[leftmargin=*,nosep]
    \item \textbf{Cross Entropy Loss (CELoss)} trains a classifier using cross-entropy loss on in-distribution documents. 
    The predicted probabilities from the classifier are used as confidence estimates for OOD detection.
    \item \textbf{Correctness Ranking Loss (CRL)}~\cite{moon2020confidence} is a regularization term added to the CE-loss to make class probabilities better confidence estimates. 
    It estimates the true class probability to be proportional to the number of times a sample is classified correctly during training.
\end{itemize}

\subsection{Experimental Settings}
We experiment with BERT~\cite{Devlin2019BERTPO} and RoBERTa~\cite{liu2019roberta} as text classifiers.
For \our, we train the classifier for 5 epochs with a batch size of 16 using an AdamW optimizer. We use a learning rate of 5e-5  using a linear scheduler with no warm-up steps.
For all baselines, we train the classifier for the same number of steps.

In our evaluation, for each dataset as in-distribution, we treat all other datasets as OOD and compute the performance.
Our evaluation follows a standard approach for each in-distribution dataset where
we begin by splitting the in-distribution dataset into three subsets: 80\% for training, 10\% for validation, and 10\% for testing. 
The model is trained using the training split, and its performance is evaluated on both the test split of the in-distribution dataset and the entire OOD dataset.

\smallsection{Evaluation Metrics} We utilize evaluation metrics from ~\cite{hendrycks2017a, devries2018learning, moon2020confidence} such as the false positive rate at 95\% true positive rate (FPR95), minimum detection error over all possible thresholds (ERR), the area under the risk-coverage curve (AURC), and the area under the precision-recall curve (AUPR) using in-distribution samples as the positives.

\subsection{Results}
We summarize the evaluation results with BERT and RoBERTa as classifiers on four datasets in Table~\ref{results-cls-all}.
All experiments are run on three random seeds and the mean performance scores are reported.
As shown in Table~\ref{results-cls-all}, we observe that \our performs better than the baselines on most of the in-distribution, OOD dataset pairs for both classifiers.
A low FPR95 value indicates that the top 95\% confident samples, selected based on their probability scores, predominantly belong to the in-distribution class and \our achieves improvements of up to 82 points in FPR95 with Yelp as in-distribution and Clima as OOD datasets when compared to CRL with BERT classifier.
\our also exhibits substantial improvements of up to 33 points in Detection Error, 48 points in AUROC, and 58 points in AUPR when compared to CE-Loss with BERT classifier.
\our achieves a perfect OOD detection score for some settings such as TREC as OOD and NYT as In-distribution datasets for both BERT and RoBERTa classifiers.
These results highlight the effectiveness of our ranking formulation with self-supervised training using IDIL loss.

\begin{table}[t]
\centering
\resizebox{\linewidth}{!}{
\scalebox{1.0}{
\begin{tabular}{l c c c c}
\toprule
Method & FPR95 & ERR & AUROC & AUPR \\
& & \begin{tabular}{@{}c@{}} In Dist: Clima \\ OOD: NYT \end{tabular} & & \\
\midrule
    \our & 17.2 & \textbf{2.6} & \textbf{97.8} & \textbf{94.5} \\
    + Grad Sub & 67.9 & 8.9 & 82.4 & 61.1 \\
    + Grad Sub \& Min & 98.4 & 10.7 & 49.2 & 32.3 \\
    + \begin{tabular}{@{}c@{}} Intra-Doc \end{tabular} & 92.7 & 10.6 & 67.3 & 41.6 \\
    - SILu & \textbf{14.6} & 3.9 & 96.7 & 89.3 \\
\midrule
& & \begin{tabular}{@{}c@{}} In Dist: NYT \\ OOD: Clima \end{tabular} & & \\
\midrule
    \our & \textbf{10.5} & \textbf{1.7} & \textbf{98.4} & \textbf{90.5} \\
    + Grad Sub only & 83.5 & 5.1 & 77.4 & 44.8 \\
    + Grad Sub \& Min & 96.5 & 5.2 & 60.3 & 36.0 \\
    + \begin{tabular}{@{}c@{}} Intra-Doc \end{tabular} & 48.3 & 2.6 & 93.2 & 80.2 \\
    - SiLU & 19.0 & 2.3 & 97.4 & 86.0 \\
            
\bottomrule
\end{tabular}					
}
}
\caption{Ablation Study. All experiments are repeated with three random seeds and the mean score is reported.}
\label{results-ablation}
\end{table}

     
            

\subsection{Ablation Study}
\label{sec:abl}
To understand the impact of each component in our IDIL loss design and implementation, we compare our method with four ablated versions with BERT classifier in Table~\ref{results-ablation}: (1) \textit{\our + Grad Sub} represents our method with backpropagating gradients through the subtrahend instead of the minuend, (2) \textit{\our + Grad Sub \& Min} represents our method with gradients backpropagating through both minuend and subtrahend, (3) \textit{\our + Intra-Doc} considers intra-document comparison similar to CE loss, in addition to inter-document intra-label comparison for loss computation, and (4) \textit{\our - SiLU} excludes SiLU function from the IDIL loss formulation.
We also present the performance of \our for reference.
\our performs better than \textit{\our + Grad Sub \& Min} demonstrating that backpropagating through one part of the difference is more beneficial than both, and the comparison between \our and \textit{\our + Grad Sub} indicates that backpropagating through minuend is better than subtrahend.
We observe that incorporating intra-document comparison into the loss formulation leads to a decrement in the ranking ability of the model.
Finally, we observe that removing the SiLU function from the IDIL loss leads to a decrease in most of the metrics.

\begin{figure*}[t]
    \subfigure[CE Loss]{
        \includegraphics[width=0.32\linewidth]{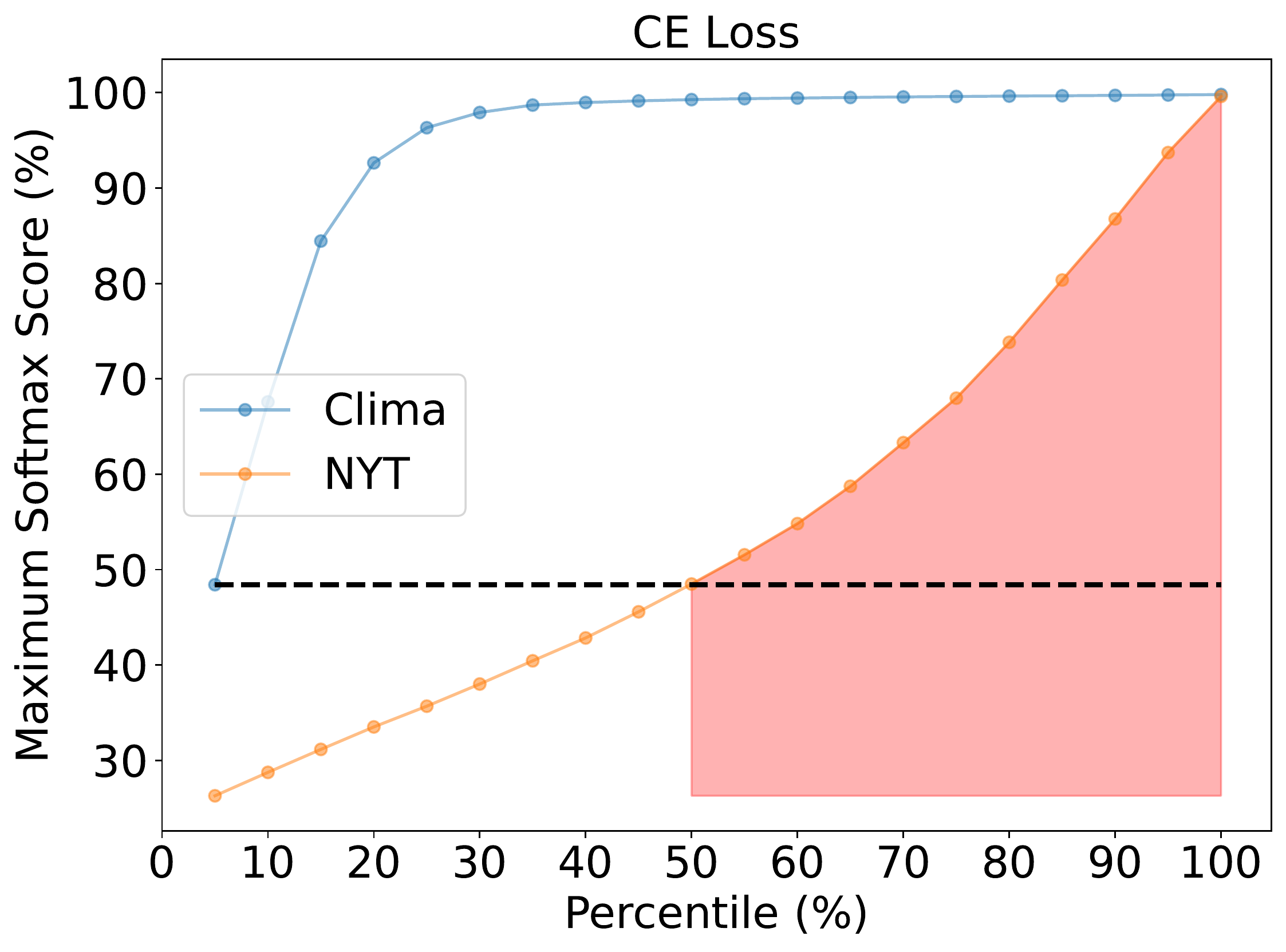}
    }
    \subfigure[CRL]{
        \includegraphics[width=0.32\linewidth]{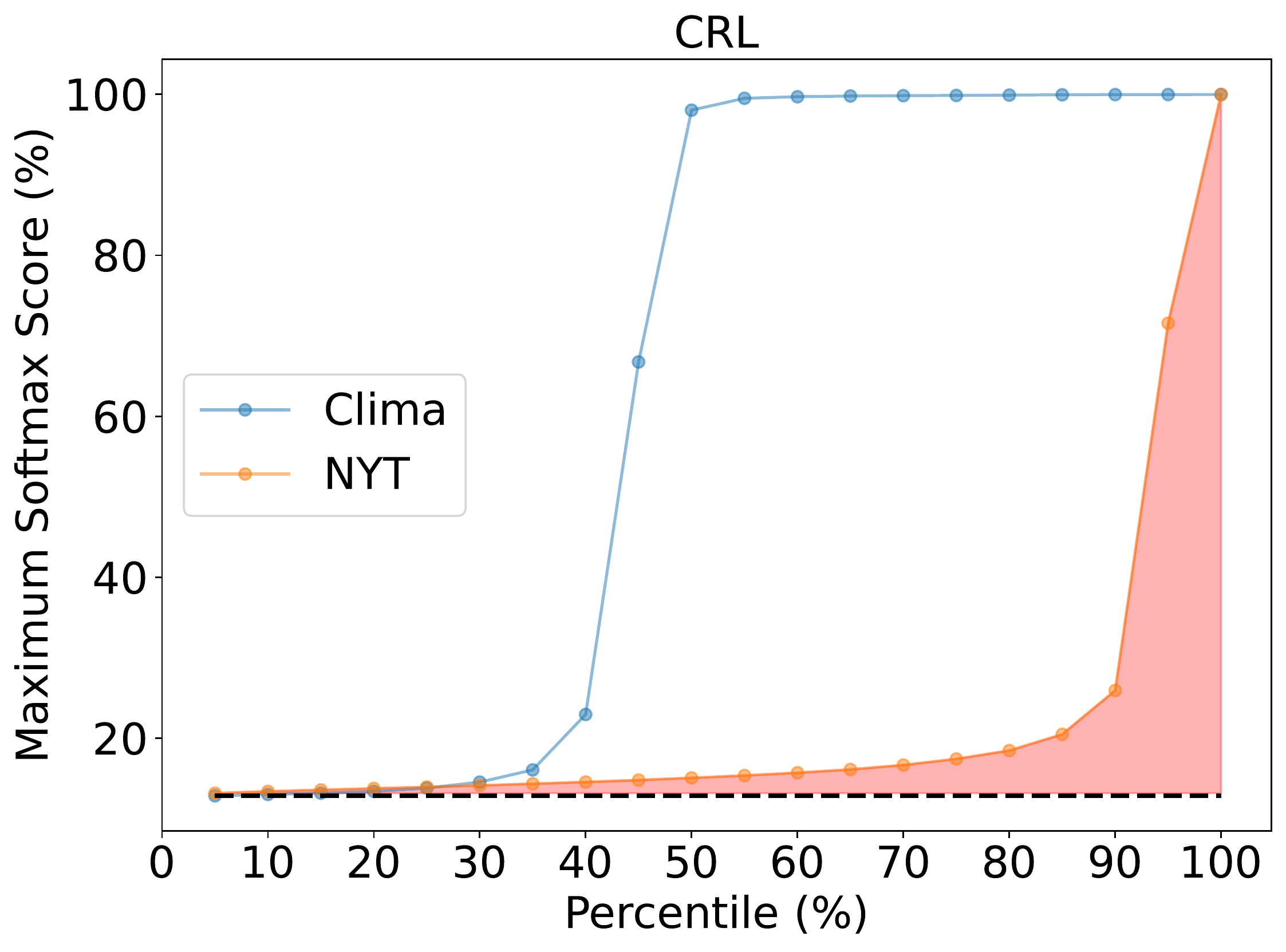}
    }
    \subfigure[\our]{
        \includegraphics[width=0.32\linewidth]{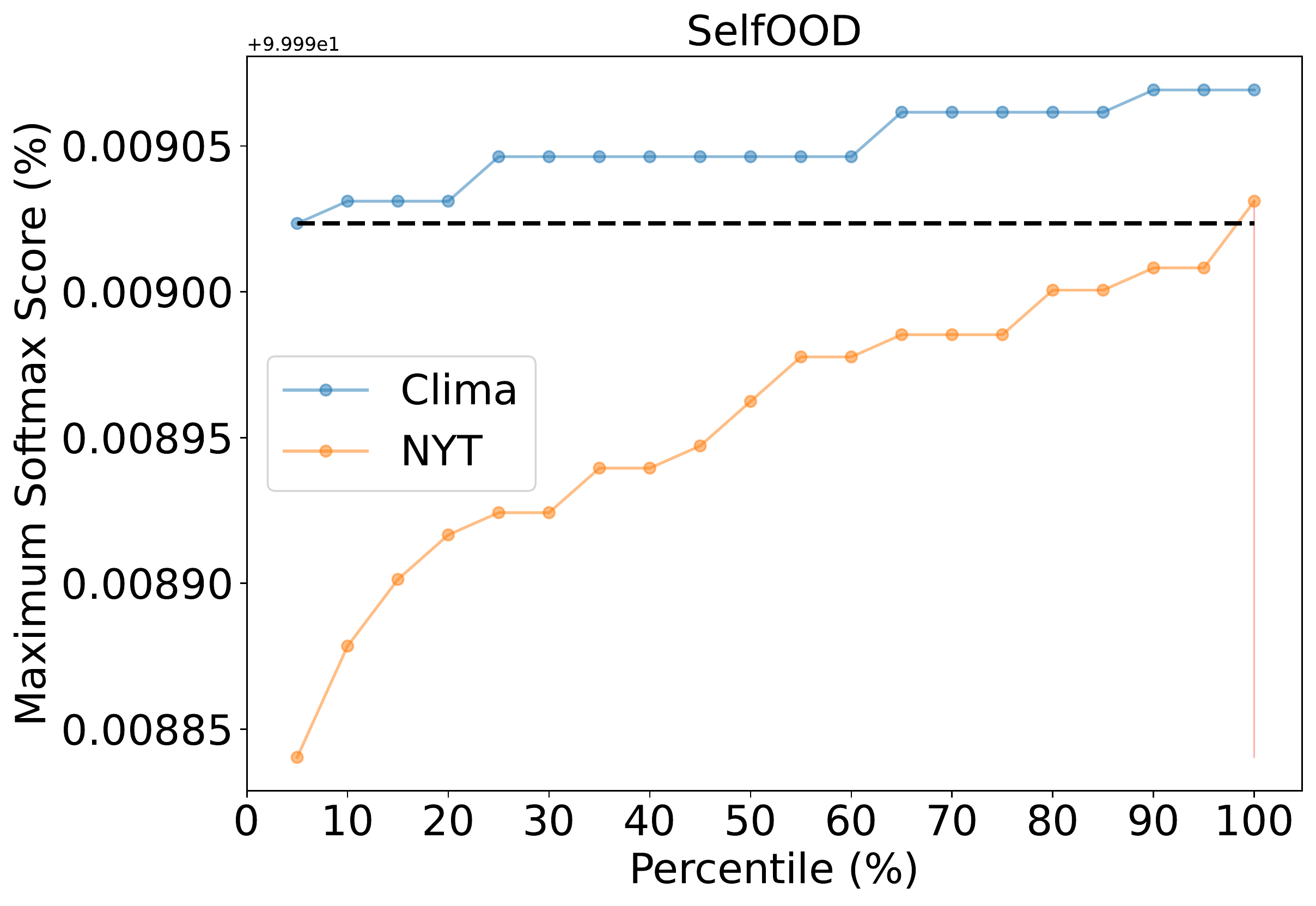}
    }
    \vspace{-3mm}
    \caption{Maximum softmax score vs percentile in-distribution \& OOD data associated with that score. By using the least in-distribution maximum softmax score as the threshold (dotted line) for OOD classification, CE loss considers more than 50\% and CRL considers almost 100\% of the OOD data as in-distribution (red region). However, in the case of \our, we observe a clear margin in maximum softmax scores that separate OOD and in-distribution.}
    \label{figure:max_softmax}
    \vspace{-5mm}
\end{figure*}

\section{Analysis \& Case Studies}
\label{sec:analysis}

In this section, we present a comprehensive analysis of our proposed method from different perspectives to understand its effectiveness.


\begin{figure}[t]
    \center
    \includegraphics[width=\linewidth]{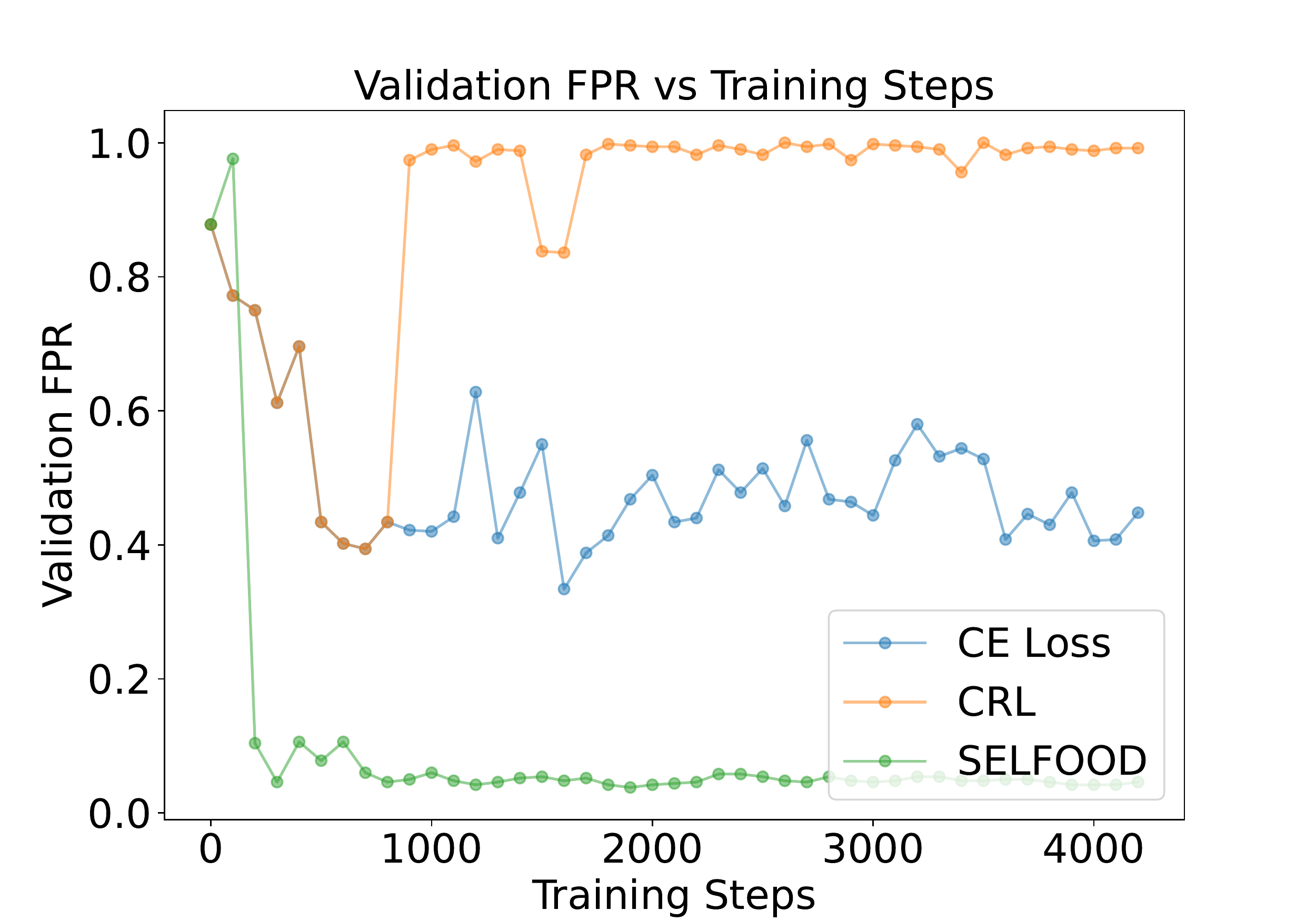}
    \caption{
    We train a BERT OOD classifier on the Clima dataset and plot validation FPR95 with Yelp as the OOD dataset during training. We observe a continuous decrease in FPR95 for \our.
    }
    \label{fig:fpr95}
\end{figure}

\subsection{Maximum softmax score Analysis}
We plot the maximum softmax score with the percentile in-distribution and OOD data associated with that score for CE loss, CRL, and \our with BERT classifier on Clima as in-distribution and NYT as OOD datasets in Figure~\ref{figure:max_softmax}.
When comparing the maximum softmax score with the percentile data associated with that score, interesting observations can be made regarding its distribution. 
Specifically, when using the least in-distribution maximum softmax score as the threshold for OOD classification, we find that the CE loss considers over 50\% of the OOD data and CRL considers almost 100\% of the OOD data as in-distribution. 
However, in the case of \our, we observe a clear margin in the maximum softmax scores that effectively separates OOD and in-distribution data. 
This suggests that the threshold needs to be carefully tuned for CE loss and CRL, requiring more effort and annotated data, whereas in the case of \our, we do not require such tuning.
This demonstrates that \our is capable of accurately classifying between OOD and in-distribution samples based on their respective maximum softmax scores, resulting in superior performance.

\subsection{Validation FPR95 vs Training steps}
We train a BERT OOD classifier on Clima as the in-distribution dataset and plot validation FPR95 vs training steps with Yelp as the OOD dataset.
As shown in Figure~\ref{fig:fpr95}, we observe that FPR95 continuously decreases and plateaus for \our whereas for other methods it fluctuates and stagnates at a higher value.
This demonstrates that the minimization of IDIL loss solely on in-distribution data successfully translates to the OOD detection task.


\subsection{Batch size Analysis}
\begin{figure}[t]
    \subfigure[NYT]{
        \includegraphics[width=0.47\linewidth]{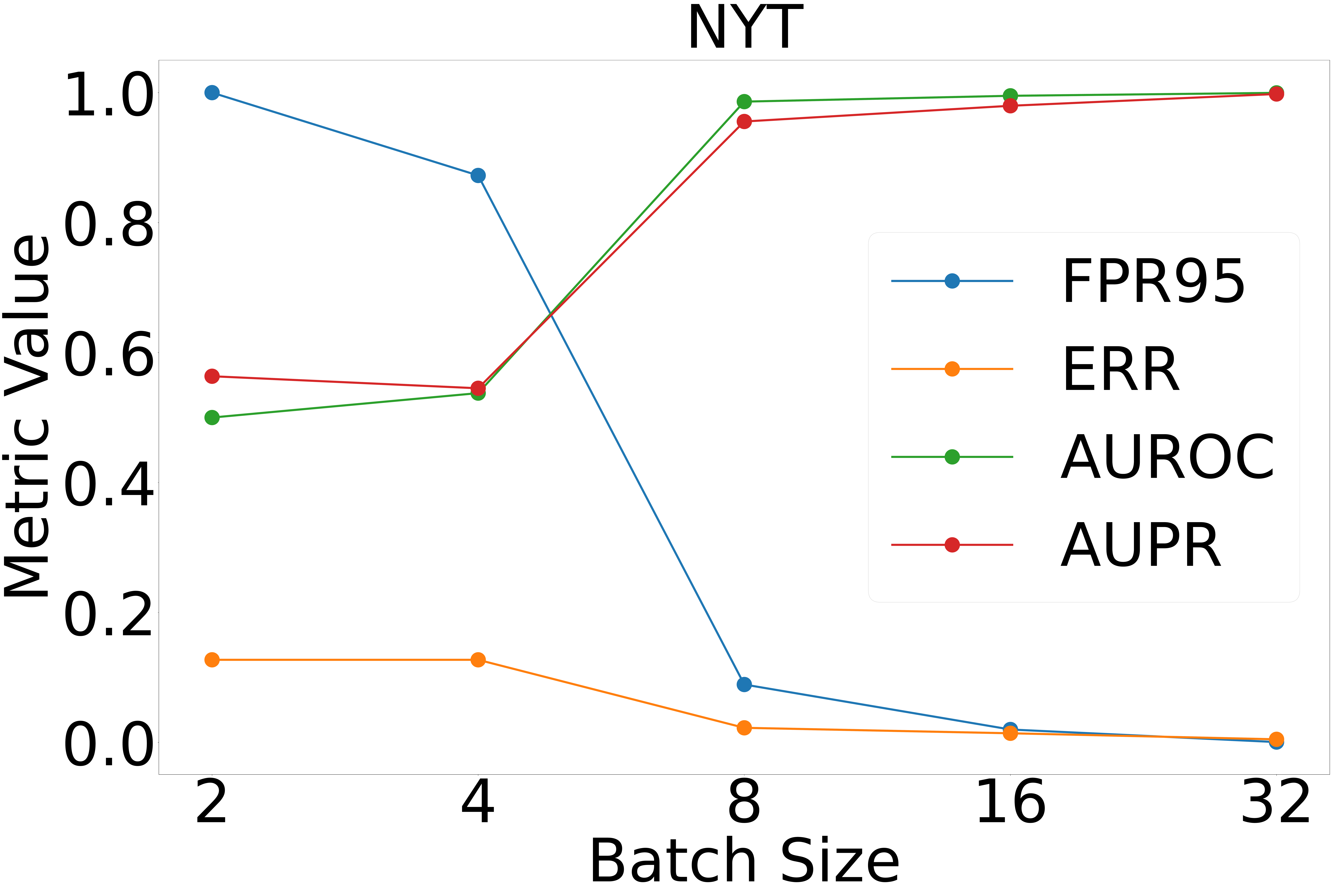}
    }
    \subfigure[Yelp]{
        \includegraphics[width=0.47\linewidth]{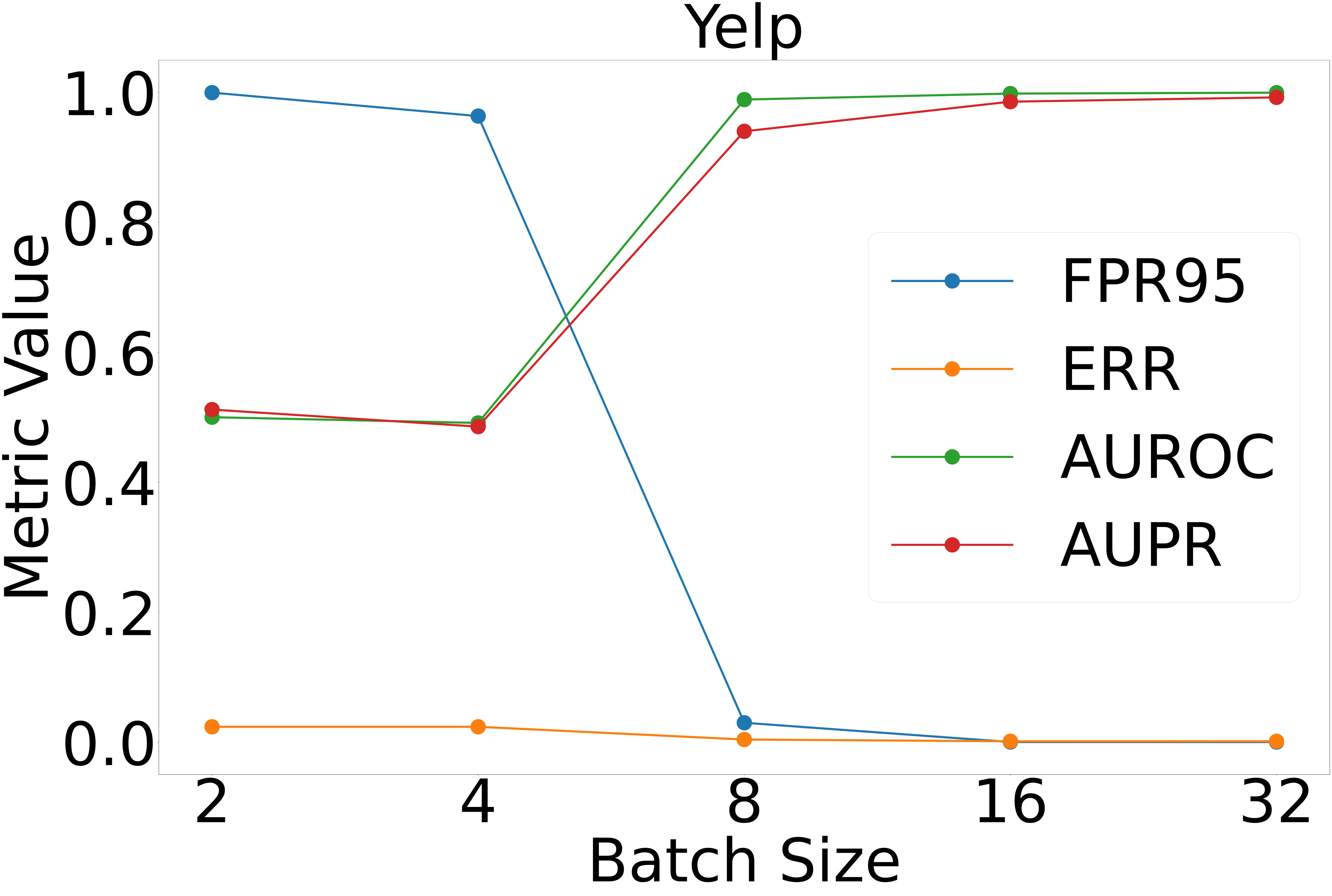}
    }
    \vspace{-3mm}
    \caption{Performance of \our w.r.t. batch size. We consider BERT classifier, Clima as in-distribution, and NYT, Yelp as OOD datasets. Each experiment is repeated with three random seeds and the mean is plotted. We observe an increase in performance with an increase in batch size.}
    \label{figure:bsize}
    \vspace{-5mm}
\end{figure}
We would like to note that in our methodology, we compare documents within the mini-batch when computing the loss. 
Consequently, the number of document pairs over which the loss is computed depends on the batch size used during training. 
To investigate the impact of batch size on performance, we vary the batch size and evaluate the corresponding performance in Figure~\ref{figure:bsize} with BERT classifier.
We consider Clima as in-distribution, and NYT, Yelp as OOD datasets.
As the batch size increases, the performance of the model also improves, up until a batch size of 16, where the performance reaches a plateau. This observation aligns with our intuition that a larger batch size allows for more in-batch document pairs to be compared, leading to more accurate loss computation.
Based on these findings, we recommend utilizing a batch size of 16 or higher to achieve optimal performance. 



\subsection{Fine-grained OOD Detection}

To explore the limits of our proposed method, we consider in-distribution and OOD datasets from the same domain.
The objective is to assess the discriminative capabilities of \our and evaluate its ability to differentiate between samples that are closely related but still OOD, which is a challenging task. 
Specifically, we consider the news domain and choose NYT as the in-distribution, and AGNews~\cite{zhang2015characterlevel}, 20News\footnote{\url{http://qwone.com/~jason/20Newsgroups/}} as OOD datasets and train BERT classifier.
As shown in Table~\ref{results-fgood}, \our performs significantly better than CE Loss and CRL on most of the metrics.
Moreover, it also achieves near-perfect scores for the AGNews as an OOD dataset, highlighting its ability to accurately identify OOD samples even when they belong to the same domain as the in-distribution dataset.

\begin{table}[t]
\centering
\small
\resizebox{\linewidth}{!}{
\scalebox{1.0}{
\begin{tabular}{c c c c c c}
\toprule
 OOD & Method & FPR95 & ERR & AUROC & AUPR \\
\midrule
\multirow{3}{*}{AGNews } & CE-Loss & 45.5 & 0.9 & 86.5 & 16.6 \\
 & \our &\textbf{0.0} & \textbf{0.1} & \textbf{100.0} & \textbf{99.4} \\
  & CRL & 77.2 & 0.9 & 80.6 & 9.5 \\
\midrule
 \multirow{3}{*}{20News } & CE-Loss & \textbf{25.4} & 6.2 & 91.9 & 39.6 \\
 & \our & 28.2 & \textbf{4.9} & \textbf{94.2} & \textbf{61.5} \\
  & CRL & 59.9 & 6.4 & 87.4 & 36.9 \\
\bottomrule
\end{tabular}					
}
}
\caption{Fine-grained OOD detection results with NYT as the in-distribution dataset with BERT classifier. We choose two datasets AGNews, 20News from the news domain, the same as NYT, and consider them as OOD datasets. The results show that \our can accurately detect OOD samples within the same domain.
}
\label{results-fgood}
\end{table}

\subsection{\our + Mahalanobis distance }
Mahalanobis distance-based estimation of OOD scores is an effective post-processing method used on trained OOD classifiers~\cite{leys2018detecting, xu2020deep}.
We investigate whether this post-processing further improves the performance of our method.
OOD scores are estimated as the distance between a test data point and the distribution of in-distribution samples using the Mahalanobis distance metric.
Following \cite{xu2020deep}, we consider the intermediate layer encodings of the OOD classifier trained on in-distribution samples as its representative distribution.
We apply this post-processing on top of \our and experiment with two in-distribution, OOD pairs with BERT classifier.
As shown in Table~\ref{results-maha}, we observe further improvement in performance, demonstrating the quality of learned representative distribution using \our.
\begin{table}[t]
\centering
\resizebox{\linewidth}{!}{
\scalebox{1.0}{
\begin{tabular}{c c c c c c c}
\toprule
InD & OOD & Method & FPR95 & ERR & AUROC & AUPR \\
\midrule
\multirow{2}{*}{Clima } & \multirow{2}{*}{Yelp } & \our & 14.7 & 0.6 & 98.0 & 90.8 \\
 & & \our + Maha&\textbf{0.8} & \textbf{0.2} & \textbf{99.2} & \textbf{97.2} \\
\midrule
 \multirow{2}{*}{Yelp } & \multirow{2}{*}{NYT } & \our & 63.2 & 19.6 & 79.4 & 82.7 \\
&  & \our + Maha & \textbf{12.4} & \textbf{4.4} & \textbf{97.4} & \textbf{97.5} \\
\bottomrule

\end{tabular}					
}
}
\caption{OOD detection results with Mahalanobis Distance post-processing technique. We choose two in-distribution, OOD pairs with BERT classifier. The results show that \our's OOD detection capabilities are enhanced with the post-processing technique.}
\label{results-maha}
\end{table}

\subsection{Classification performance of \our}
We present the text classification performance of \our and CE-Loss with BERT classifier on in-distribution datasets in Table~\ref{results-clsperf}. 
The findings highlight that while \our serves as a reliable OOD detector, its performance as a text classifier is subpar.
This observation can be attributed to our IDIL loss formulation, which focuses on comparing confidence levels across documents for each label rather than across labels for each document. 
As a result, the IDIL loss primarily promotes the ordinal ranking of confidence levels across documents, which enhances the model's OOD detection capabilities. 
However, this emphasis on inter-document ranking comes at the expense of inter-label ranking, resulting in limited classification capabilities.
Moreover, when we introduce the intra-document comparison to the IDIL loss, as discussed in Section~\ref{sec:abl}, we observe a decline in the model's ranking ability. 
This further supports the notion that balancing the inter-document and intra-document comparisons is crucial for achieving optimal performance in both OOD detection and text classification tasks.


\begin{table}[t]
\centering
\small
\scalebox{1.0}{
\begin{tabular}{c c c}
\toprule
Dataset & Method & Accuracy \\
\midrule
\multirow{2}{*}{NYT } & CE-Loss & \textbf{97.4} \\
 & \our & 0.3 \\
\midrule
\multirow{2}{*}{Clima} & CE-Loss & \textbf{85.1} \\
 & \our & 9.5 \\
\midrule
\multirow{2}{*}{Yelp} & CE-Loss & \textbf{69.5} \\
 & \our & 20.0\\
\midrule
\multirow{2}{*}{TREC} & CE-Loss & \textbf{97.1} \\
 & \our & 1.8\\
\bottomrule
\end{tabular}					
}
\caption{Classification Performance of \our and CE-Loss with BERT classifier on two datasets. The results show that \our is a poor text classifier.}
\label{results-clsperf}
\end{table}
\section{Conclusion}

In this paper, we present \our, a novel framework for OOD detection that leverages only in-distribution samples as supervision. 
Building upon the insight that OOD samples typically do not belong to any in-distribution labels, we formulate the OOD detection problem as an inter-document intra-label ranking task. 
To address this challenge, we propose a novel IDIL loss, which guides the training process.
Through extensive experiments on multiple datasets, we demonstrate the effectiveness of \our on OOD detection task. 
However, we also acknowledge that it comes at the expense of text classification performance.
Future research can focus on developing techniques that effectively balance inter-document and intra-document comparisons, enabling improved performance in both OOD detection and text classification tasks.



\typeout{}
\bibliography{custom}
\bibliographystyle{acl_natbib}

\newpage
\appendix


\end{document}